\documentclass[sigconf,nonacm]{acmart}

\renewcommand\footnotetextcopyrightpermission[1]{} 

\AtBeginDocument{%
  }

\copyrightyear{2025}
\acmYear{2025}
\setcopyright{acmlicensed}\acmConference[MMFood '25]{Proceedings of the 1st International Workshop on Multi-modal Food Computing}{October 27--28, 2025}{Dublin, Ireland}
\acmBooktitle{Proceedings of the 1st International Workshop on Multi-modal Food Computing (MMFood '25), October 27--28, 2025, Dublin, Ireland}
\acmDOI{10.1145/3746264.3760496}
\acmISBN{979-8-4007-2046-8/2025/10}

\begin{document}

\title{Extending FKG.in: Towards a Food Claim Traceability Network}

\author{Saransh Kumar Gupta}
\authornote{These authors contributed equally.}
\authornote{Corresponding Authors}
\email{saransh.gupta@ashoka.edu.in}
\orcid{0009-0000-5887-2301}

\author{Rizwan Gulzar Mir}
\email{rizwan.mir_phd25@ashoka.edu.in}
\orcid{0009-0005-1347-2239}
\authornotemark[1]

\author{Lipika Dey}
\email{lipika.dey@ashoka.edu.in}
\orcid{0000-0003-3831-5545}
\authornotemark[2]

\author{Partha Pratim Das}
\email{partha.das@ashoka.edu.in}
\orcid{0000-0003-1435-6051}

\author{Anirban Sen}
\email{anirban.sen@ashoka.edu.in}
\orcid{0000-0002-2337-7957}
\authornotemark[2]

\affiliation{%
  \institution{Ashoka University}
  \city{Sonepat}
  \state{Haryana}
  \country{India}
}

\author{Ramesh Jain}
\affiliation{%
  \institution{UCI Institute for Future Health}
  \city{Irvine}
  \state{California}
  \country{USA}}
\email{jain@ics.uci.edu}
\orcid{0000-0003-2373-4966}

\renewcommand{\shortauthors}{Gupta et al.}

\begin{abstract}
  The global food landscape is rife with scientific, cultural, and commercial claims about what foods are, what they do, what they should not do, or should not do. These range from rigorously studied health benefits (probiotics improve gut health) and misrepresentations (soaked almonds make one smarter) to vague promises (superfoods boost immunity) and culturally rooted beliefs (cold foods cause coughs). Despite their widespread influence, the infrastructure for tracing, verifying, and contextualizing these claims remains fragmented and underdeveloped. In this paper, we propose a \textit{Food Claim-Traceability Network} (FCN) as an extension of FKG.in, a knowledge graph of Indian food that we have been incrementally building. We also present the ontology design and the semi-automated knowledge curation workflow that we used to develop a proof of concept of FKG.in-FCN using Reddit data and Large Language Models. FCN integrates curated data inputs, structured schemas, and provenance-aware pipelines for food-related claim extraction and validation. While directly linked to the Indian food knowledge graph as an application, our methodology remains application-agnostic and adaptable to other geographic, culinary, or regulatory settings. By modeling food claims and their traceability in a structured, verifiable, and explainable way, we aim to contribute to more transparent and accountable food knowledge ecosystems, supporting researchers, policymakers, and most importantly, everyday consumers in navigating a world saturated with dietary assertions.
\end{abstract}



\maketitle

\section{Introduction}
\label{sec:introduction}

\par Food is not merely material sustenance but a dense web of beliefs, narratives, and knowledge claims. Across cultures, people assert what food does, where it comes from, how it should be prepared or combined, what it symbolizes, what health benefits it has, and how it ought to be consumed, among many others. Such claims may stem from biochemistry, religious doctrine, sensory judgment, or moral conviction. Some appear in peer-reviewed journals, but most others circulate through kitchens, WhatsApp forwards, online forums, or institutional guidelines. Mundane or radical, local or global, food-related claims shape how societies understand nutrition, tradition, value, and identity. Recent decades have witnessed an unprecedented explosion of food information - from age‑old lore passed through generations to slick, data‑driven advertising campaigns by multinational brands. While some claims rest on robust science, many persist through cultural resonance, anecdote, or marketing spin. The result is an epistemic fog and a tangled ecosystem where consumers, researchers, and policymakers alike struggle to distinguish evidence‑based guidance from myth, exaggeration, or misinformation.

\par Despite their ubiquity, diversity, and significance, food-related claims remain underrepresented and untraceable in knowledge infrastructures. Nutrition databases capture quantifiable facts (e.g., macronutrients, portion sizes) but rarely qualitative or contextual claims like "turmeric heals inflammation," "microwaves destroy nutrients," or "cold water aids digestion." Worse still, these claims mutate across modalities - scientific papers, cultural discourse, advertising, digital media (e.g., health influencers and vernacular YouTube channels) - losing nuance, misrepresenting facts, or gaining unwarranted authority. Additionally, falsehoods like "raw garlic cures cancer" spread faster than claim corrections, while even valid claims (e.g., "fermented foods support gut health") often lack traceable provenance. The stakes are high because food claims shape public perceptions of health, influence agricultural practices, and frame the ethics of global supply chains. Yet the underlying infrastructure to systematically capture, validate, and trace each food claim across its multiple manifestations 
remains largely fragmented at best and nonexistent at worst.

\par In this paper, we propose the Food Claim-Traceability Network (FCN) and its knowledge curation workflow: a unified framework for modeling food claims as knowledge artifacts and tracking their origins, transformations, and evidence across domains, formats, and modalities, focusing on Indian food. FCN is being developed as an interoperable extension to FKG.in \cite{Gupta2024FKG, Gupta2025FKG}, a broader Knowledge Graph for Indian food modeling recipes, ingredients, and nutrition. While FKG.in focuses on tangible culinary entities, the FCN models claims as first-class objects - semantic assertions (e.g., "cumin aids digestion," "caffeine prevents Parkinson's disease," "walnuts improve memory") situated in time, context, and community - and links each claim to relevant food items in FKG.in. The framework enables food claims to be extracted, parsed, contextualized, structured, linked to evidence, and integrated into larger culinary knowledge systems. With advances in large language models (LLMs), this can be semi-automated at scale across diverse sources: Reddit, scientific papers, blogs, packaging, and folk discourse. This paper outlines the first steps in this direction. We present our modular system design, a lightweight ontology for food claim representation, and a proof-of-concept claim extraction and validation pipeline using a small Reddit dataset. While the current prototype is not yet grounded in Indian food claims, it is designed to be adaptable and will be further refined to suit the Indian context. At the same time, it remains extensible to other geographic, culinary, and regulatory ecosystems.

\par The rest of the paper is organized as follows. Section \ref{sec:related_work} reviews existing work on food ontologies, knowledge graphs, misinformation, and scientific claim verification. Section \ref{sec:motivation} discusses the complexities and urgency of the dietary claim-related problems in the Indian context. Section \ref{sec:food_claims} covers the scope, types, and challenges of studying and structuring food claims. Section \ref{sec:methodology} details the ontology design, semi-automated knowledge curation workflow for FKG.in-FCN, and a proof-of-concept demonstration using Reddit data. Section \ref{sec:results} presents analytical results and interpretations. Section \ref{sec:limitations} discusses limitations and future directions. Section \ref{sec:conclusion} summarizes key contributions, reflects on FKG.in-FCN's impact, and emphasizes its potential as a foundational resource for future research in food claim validation and misinformation mitigation.

\section{Related Work}
\label{sec:related_work}

\par Recent research has extensively explored ontologies to formalize, integrate, and enhance reasoning over food-related data across domains. FoodOn \cite{Dooley18} represents the farm-to-fork lifecycle, unifying food terminology - including ingredients, preparation methods, and packaging metadata - and ensuring interoperability with vocabularies like LanguaL \cite{Moller2017LanguaL}. ISO-FOOD \cite{Eftimov2019ISOFOOD} focuses on isotopic and provenance metadata in food science. Ontologies like ONS \cite{Vitali2018ONS} and FOBI \cite{Castellano2020FOBI} extend this scope into nutrition and metabolomics: ONS standardizes dietary study terminology, integrating CHEBI \cite{Degtyarenko2007ChEBI}, FoodOn, and OBI \cite{Bandrowski2016OBI}, and adds specialized terms for biomarkers and dietary protocols, while FOBI links food intake data with biomarkers for metabolic tracking. Domain-specific efforts, such as the Seafood Ontology \cite{Vinu2018}, classify seafood based on quality assurance protocols in Protégé \cite{musen2015protege}. These ontologies provide semantic rigor and data harmonization but primarily focus on food classification, nutrition, and supply-chain provenance, rather than discourse-level reasoning or claims.

\par Parallel research has applied knowledge graphs to improve interpretability, reasoning, and personalized digital health \cite{iso_11147_2023}. FoodKG \cite{Haussmann19} enables nutrient-aware recipe search and Q\&A. RcpKG \cite{lei2021} incorporates multimodal data - user preferences and food images - via graph convolution and BERT-based models for recommendations. The World Food Atlas Project \cite{rostami2021} aims to create a comprehensive food knowledge graph by linking ingredients, cultures, and dietary habits across countries, while AgriKG \cite{chen2019} aims for agricultural and cultural coverage by extracting large-scale unstructured food knowledge via NLP and deep learning. In health, dietary knowledge graphs connect food intake and nutrition to symptoms, diseases, and population factors \cite{lan2019, Milanlouei2020}, and personalized dietary guidance systems like \cite{akilesh2025} use LLMs with Neo4j-based graphs for tailored diets. FKG.in \cite{Gupta2024FKG, Gupta2025FKG} builds a structured graph of Indian culinary data, linking recipes, ingredients, nutrition, and food practices using interoperable ontologies and LLMs inspired by FoodOn and FoodKG, aiming to support dietary analysis, health recommendation, cultural preservation, and nutrition informatics. Collectively, these works demonstrate food knowledge graphs' versatility and growing significance in powering next-generation dietary guidance, health monitoring, and culturally-aware food applications.

\par In contrast, little work has modeled food-related claims, particularly regarding misinformation, beliefs, and traceability. Scientific claim verification frameworks include SciFact \cite{wadden2020fact, wadden2021overview, wadden2022multivers, wadden2022scifact}, which provides benchmarks, shared tasks, and methods leveraging document context and weak supervision, and schema-based approaches like VERISCI \cite{Ze2023} and FEVER-style frameworks \cite{thorne2018fever}. For food, the Journal of Nutrition highlights growing misinformation and calls for frameworks supporting critical and ethical thinking \cite{diekman2023misinformation}, echoing concerns from the American Dietetic Association \cite{ayoob2002position} and other scholarly and public health literature \cite{nestle2002food, feldman2005panic, pollan2008defense, nestle2018unsavory, wolrich2021food, vantulleken2023ultra}. Numerous studies document nutrition misinformation on social media \cite{Rodrigues2024, Diyab2025NutritionMisinformation, thompson2016carrots}, yet most do not systematically track provenance. Our work addresses this gap by proposing a food claim-centric ontology and graph-based network to capture semantic structure, claim origins, and validation sources, combining claim-verification methods with domain-specific knowledge representations to address food misinformation and interpretive complexity.

\section{Food Claims in India: Urgency and Impact}
\label{sec:motivation}

\par India presents a uniquely complex challenge for food claim verification due to its linguistic diversity, cultural plurality, and widespread advertising and digital misinformation. Food beliefs in India are deeply entangled with language, region, caste, ritual, and traditional medicine. From WhatsApp forwards about turmeric cures to YouTube videos promoting miracle diets, consumers navigate an overwhelming and often contradictory web of information, with little means to trace, verify, or contest claims. Regional contradictions and culturally embedded practices further complicate matters: rice is prescribed during illness in Tamil Nadu and West Bengal, where preparations like \textit{kanji} (fermented rice water) or \textit{payesh} (rice pudding) are considered easy on stomach and restorative but avoided in Uttar Pradesh or Rajasthan during cold due to the belief that it cools the body or increases phlegm. These opposing claims arise from regional traditions, climate logic, and local medicine systems rather than intrinsic nutritional properties of rice. Without structured traceability, such contradictions coexist unchecked, preventing critical engagement or evidence-based comparison.

\par Low health literacy and limited English proficiency in India exacerbate the problem: most consumers ignore nutrition and ingredient labels or disclaimers on packaged food \cite{Saha2013, shireen2022food, ali2009understanding, kumar2017do, Vemula2014}. Instead, their beliefs are shaped by oral transmission, advertising campaigns, cultural authority, or viral digital content. The widespread adoption of smartphones and social media - 530M WhatsApp, 516M Instagram, 295M X, and 465M YouTube users \cite{theglobalstatistics2025, theglobalstatistics2025youtube} - has amplified the issue, making India a fertile ground for food myths and misinformation. Compounded by vernacularized food facts, selective scientific language to legitimize traditional beliefs, lack of critical media literacy, and limited claim traceability, the result is an information landscape where myths flourish, while credible corrections rarely catch up. A popular example is the belief that carrots improve eyesight - originally spread as World War II propaganda \cite{thompson2016carrots} - which remains widely believed in India despite being debunked.

\par Beyond cultural contradictions and linguistic barriers, the urgency for structured food claim traceability in India is amplified by public health risks, regulatory gaps, and skewed food hierarchies. Fad diets imported from Western wellness culture, often promoted by influencers, highlight exotic ingredients like quinoa or avocado while ignoring indigenous foods with comparable nutrition. E.g., amaranth - an ancient Indian grain rich in protein, iron, and calcium - is overlooked despite its availability and affordability, while quinoa is hailed as a superfood in urban diets. These trends alienate local food systems and reinforce aspirational consumption tied to class and status. Limited claim-level traceability also hampers regulatory bodies like FSSAI (Food Safety and Standards Authority of India) from verifying nutritional or medicinal assertions on packaging or advertisements, allowing misleading health halos to persist unchecked, which can promote unhealthy nutritional practices like 'raw-food', 'crash' or 'alkaline' diets, especially among youth. Without multilingual, culturally grounded, evidence-linked infrastructures for claim validation, India remains vulnerable to both imported myths as well as locally entrenched half-truths.

\par Together, these factors make India not only vulnerable to food misinformation but also an urgent testbed for designing traceable, structured, and culturally sensitive food knowledge systems.

\section{Food Claims: Overview, Scope, and Challenges}
\label{sec:food_claims}

\par Food claims assert a relationship between a food item and some property, effect, or outcome - health-related, cultural, moral, etc. Unlike quantitative nutritional data, qualitative claims are context-dependent, operating in the realm of beliefs, intentions, and authority. Their challenge lies not just in variability, but in their embedding in diverse communicative and complex epistemic settings. The Food Claim Network (FCN) aims not merely to identify claims but to systematically structure, contextualize, and trace them. Currently, we focus on claims tied to specific foods (e.g., soaked almonds make one smarter) rather than broader claims (e.g., locally sourced food lowers carbon footprint) to enable efficacy and easier linking to entities in FKG.in. Our scope extends beyond health or commercial labeling to model claims across domains, as exemplified in Table \ref{tab:claim_types}.

\begin{table*}
    \caption{Different Types and Examples of Food and Health Claims Common in India}
    \label{tab:claim_types}
    \begin{tabular}{p{0.1cm}p{3.1cm}p{6.4cm}p{6.4cm}} \toprule
    \multicolumn{1}{c}{\textbf{No.}} & \multicolumn{1}{c}{\textbf{Claim Type}} & \multicolumn{1}{c}{\textbf{Claim Example 1}} & \multicolumn{1}{c}{\textbf{Claim Example 2}} \\ \midrule 
    1. & Scientific/Medical & White rice (high GI) spikes insulin levels. & Curcumin inhibits inflammatory pathways. \\ 
    2. & Cultural/Traditional & Eating curd at night causes a cold. & Spicy foods cause heat in the body. \\ 
    3. & Moral/Political & Chicken \textit{tikka} from factory poultry is unhealthy. & Eating beef violates \textit{dharmic} (virtuous) principles. \\ 
    4. & Sustainability/Regulatory & Millets require less water than wheat. & Fortified flour meets national nutrition guidelines.\\ 
    5. & Aesthetic/Sensory & Aged cheddar cheese has a more desirable flavor. & Mango contains more \textit{prana} (energy) than juice. \\ 
    6. & Religious/Ritualistic & Garlic and onion are \textit{tamasic} (dullness-inducing). & \textit{Halal} chicken is permissible in Islamic festivals. \\ 
    7. & Social/Symbolic & Tofu is a healthier and ethical substitute for paneer. & Millets are ancient food, not fit for modern diets. \\ 
    8. & Origin/Authenticity & \textit{Basmati} is the aromatic rice variety. & \textit{Goan vindaloo} must use toddy for authentic flavor. \\ 
    9. & Marketing/Advertising & Almond milk contains no cholesterol. & Cold-pressed beetroot juice detoxifies the liver.\\ 
    10. & Digital/Influencer & Avocado toast is the perfect food for productivity. & Fasting with bulletproof coffee resets gut. \\ \bottomrule \end{tabular} \end{table*}

\par Key challenges in studying food claims, notably in India, include:
\begin{itemize}
    \item \underline{Ambiguity and multi-intentionality}: The same claim can imply different things across contexts. E.g., "rice is heavy" may refer to poor digestibility in Ayurveda, high caloric density nutritionally, or unsuitability during hot weather regionally.
    \item \underline{Source and transformation tracking}: Claims mutate across scientific, cultural, or marketing formats, often losing precision or exaggerated. E.g., a clinical finding on turmeric's anti-inflammatory compounds may become "turmeric cures all inflammation" in advertising.
    \item \underline{Validation mismatch}: Some claims resist biochemical and medical validation but remain meaningful in experiential, cultural, or ecological logics. E.g., "homemade tomato soup helps relieve a cold" may lack clinical trials, but is validated across generations.
    \item \underline{Scope explosion}: Claims extend beyond health or branding to encode identity, purity, cultural belonging, symbolism, morality, accessibility, and emotion. A single claim may serve as a dietary tip, a class signal, a political stance, and a spiritual assertion simultaneously, complicating efforts to validate or contest it in isolation. E.g., \textit{Khichdi} (rice-lentil porridge) without onion and garlic is \textit{sattvic} (pure) and calming.
    \item \underline{Contradictory nature}: Many claims evolve and conflict over time. E.g., egg yolks were long vilified for cholesterol but later recognized for nutritional density and limited heart impact. Such shifts in consensus highlight the need for structured traceability to correct misinformation as well as to preserve historical, cultural, and scientific context.
\end{itemize}
\par To address these challenges, we ground claims in a structured ontology, providing a scaffold for their comparative, cross-contextual, and historical study at scale, laying foundation for computational food semiotics and accountable food knowledge infrastructures.

\section{Methodology}
\label{sec:methodology}

\subsection{Proposed Ontology Design for FKG.in-FCN}
\label{subsec:ontology_design}

\begin{figure*}[t]
  \centering
  \includegraphics[width=\textwidth, trim=0 20 0 0, clip]{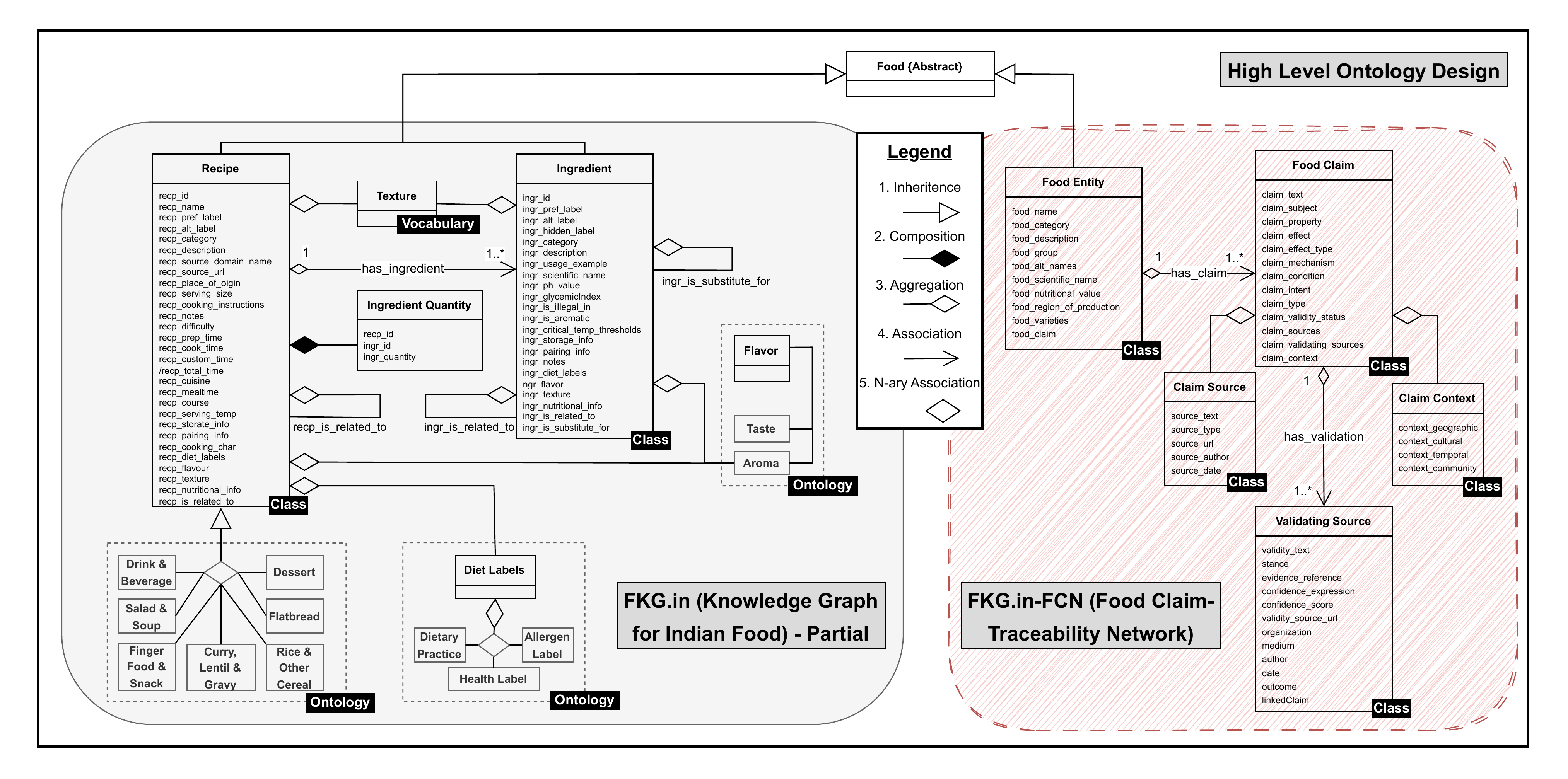}
  \caption{High Level Ontology Design of FCN as an extension to FKG.in} 
  \label{fig:fcn_ontology}
  \Description{Overview of the Food Claim Ontology. The ontology models food-related claims as structured, traceable entities. At its core is the Claim class, representing a distinct assertion about food, linked to its originating ClaimSource and optionally situated within a ClaimContext (e.g., geographic or cultural setting). Claims are further linked to one or more ValidatingSource instances, which capture supporting, opposing, or clarifying responses across modalities, such as scientific studies, cultural commentary, or online discussions. The ontology is designed to support multimodal provenance, stance tracking, and cross-contextual reasoning, enabling fine-grained analysis of how claims propagate and transform across knowledge systems.}
\end{figure*}

\par In this section, we describe the ontological components of FCN. While motivated by the studies in Section \ref{sec:related_work}, our ontology addresses gaps in existing literature by focusing on food-related misconceptions, misinformation, and controversies. This is crucial given the overload of food information from multiple sources and modalities, where consumers struggle to identify a single credible source. The proposed ontology aids in verifying claim provenance, ensuring validation and transparency for informed decision-making. By explicitly modeling the entities involved in food claims and their intents, the ontology enables structured reasoning over conflicting or dubious information, supporting tasks like claim verification, source credibility assessment, and automated misinformation detection. Unlike existing ontologies, it emphasizes the information ecosystem surrounding food discourse, not nutritional content or food categorization.

\par In \cite{Gupta2024FKG}, we conceptualized a general category, \textbf{Food}, with subcategories, \textbf{Recipe} and \textbf{Ingredient}, capturing commonalities while preserving the respective distinctions of the subcategories for downstream applications. Continuing this approach, we now add a third subcategory, \textbf{Food Entity}, and plan to incorporate elements from the ClaimReview schema \cite{schemaClaimReview} for structured representation and interoperability. Figure \ref{fig:fcn_ontology} shows the proposed FCN ontology with a partial FKG.in ontology for reference. The major categories introduced in FCN are described below:
\begin{enumerate}
    \item \textbf{Food Entity}: The main food entity mentioned in a claim may be a dish/recipe or an individual ingredient. A \textbf{Food Entity} instance has all the properties of either the Recipe or Ingredient categories, depending on its classification. We also include additionally important properties like \textbf{category}, \textbf{description}, \textbf{group}, \textbf{alternate names}, \textbf{scientific name}, \textbf{nutritional value},\textbf{ regions of production}, and \textbf{varieties} in \textbf{Food Entity}. In prompt engineering experiments, we observed that source data often includes these details along with claims, which we extract simultaneously using an LLM. Thus, we obtain 2 sets of values for these properties: one set is extracted by the LLM strictly from the input text, while the other is inferred by the LLM independently, without relying on the text. This distinguishes explicitly stated from contextually inferred information, enabling a nuanced representation of \textbf{Food Entity}. It also facilitates source traceability, reasoning-based augmentation, confidence scoring, human-in-the-loop validation, and adaptive refinement of food-related ontologies in diverse cultural and linguistic contexts. Occasionally, a \textbf{Food Entity} instance may turn out to be a meal like "Breakfast", or a nutrient such as "Vitamin B" when the input text does not refer to a specific food item, leaving most of the above-mentioned properties empty.
    \item \textbf{Food Claim}: In this core category of  FCN, each food claim is represented as a structured object containing semantic and contextual components that help in interpretation, comparison, and validation. Below, we describe the different grammatical components of the \textbf{Food Claim} instance:
    \begin{itemize}
        \item \textbf{claim\_text}: The original sentence or phrase from which the claim is extracted. 
        \item \textbf{claim\_subject}: The primary and singular entity that the claim is about, resolving to a specific instance of \textbf{Food Entity}, \textbf{Ingredient}, or \textbf{Recipe} categories, such as "dragon fruit". Occasionally, it may also refer to a nutrient or compound, such as "Vitamin C in orange". 
        \item \textbf{claim\_property}: A qualitative or quantitative attribute or state of the subject being described. This differs from and often precedes the effect, as it describes what the subject has (characteristic) rather than what it does (an action verb), such as  "rich in antioxidants" or "low in cholesterol".
        \item \textbf{claim\_effect}: The action, change, or influence the subject (e.g., food, supplement, herb, nutrient) is claimed to cause in the body or mind, including physiological (e.g., "lowers blood pressure"), perceptual (e.g., "makes skin glow"), disease-related (e.g., "prevents cancer"), cognitive (e.g., "improves memory"), cultural (e.g., "cools the body") or functional (e.g., "aids digestion", "boosts energy") effects.
        \item \textbf{claim\_effect\_type}: High-level category of the effect to aid filtering, clustering, and analyzing claims. Multiple effect types may apply to a single claim. Examples include "health", "mental", "skin", "metabolism", "detox", "energy", "weight", "immunity", "digestion", etc.
        \item \textbf{claim\_mechanism}: Causal pathway or biological explanation behind how the effect occurs, typically starting with expressions like "by", "via", or "through". E.g., "by binding free radicals", "through enhancing insulin sensitivity", etc.
        \item \textbf{claim\_condition}: Temporal, contextual, or situational qualifiers that enable, limit, or modify the claim's applicability. E.g., "on an empty stomach", "if consumed daily".
        \item \textbf{claim\_intent}: Singular epistemic classification based on truth-value and intent: "fact", "myth", "misrepresentation", "misinformation", "disinformation", or "malinformation".
        \item \textbf{claim\_type}: Captures the conceptual nature of the claim for organization and comparison across domains. Multiple claim types may apply to a single claim, with the value being the types identified in Table \ref{tab:claim_types}.
        \item \textbf{claim\_validity\_status}: Coarse-grained judgment about the scientific validity of the claim, marked as strictly one of: "true" (if backed by a source), "false" (if debunked), or unverified (e.g., traditional beliefs with no evidence).
    \end{itemize}
    \item \textbf{Claim Source}: This category captures the origin or initial reporting of a food claim, such as a research article, newspaper, blog post, regulatory document, social media post, etc. It anchors each claim to a tangible source, allowing us to track how claims first enter the public domain or information ecosystem. While some claims originate in formal publications or official channels, many emerge in informal or in culturally embedded ways. Thus, this category accommodates diverse source types with flexible metadata to describe a claim's first appearance or point of capture. This foundational anchoring enables downstream analysis of claim propagation, mutation, or contestation across contexts.
    \item \textbf{Claim Context}: This category provides semantic grounding for a claim by situating it in its geographic, cultural, temporal, or epistemic setting in which it is typically made or believed. For instance, "curd should not be eaten at night" is prevalent in parts of South Asia, while "milk and fish should not be eaten together" spans multiple cultures with varying justifications. Geographic and sociocultural tagging disambiguates similar claims with different foundations, supporting comparative studies on regional variation, cultural specificity, or localization of health and food beliefs. By modeling where, when, and in what context a claim is situated or most salient, the \textbf{Claim Context} layer enables richer reasoning about credibility, relevance, and stakeholder perspectives.
    \item \textbf{Validating Source}: This category captures the web of commentary and scrutiny that surrounds a food claim after it enters discourse. It includes external sources that support, challenge, request evidence for, question, or clarify a given claim, whether from scientific studies, expert statements, regulatory comments, anecdotal testimonies, or online discussions. Each validation entry records stance, medium, speaker, and source type, allowing claims to be contextualized within diverse knowledge communities and preserving multiple perspectives. E.g., a claim may be simultaneously challenged by a scientific review, supported by a traditional knowledge, and hedged by consumer anecdotes - all of which are stored as linked but independent viewpoints. Thus, each claim may have more than one validating source. The inclusion of confidence indicators, provenance metadata, and stance classification enables users to trace how different types of evidence and authority interact around a claim. Unlike \textbf{Claim Source}, which records where a claim first appears, \textbf{Validating Source} traces how it is subsequently evaluated, interpreted, or contested. The overarching aim is to treat validation not as a binary truth check but as a layered ecosystem of commentary, traceability, credibility, and source fidelity.
\end{enumerate}

\par Ontology specifications and supplementary resources are available on GitHub\footnote{\label{fn:repo}\url{https://github.com/fkg-india/foodcomputing-ashoka-resources}} and archived on Zenodo for long-term access\cite{fkg-india_foodcomputing_resources}.

\subsection{FKG.in-FCN Knowledge Curation Workflow}
\label{subsec:knowledge_curation}

\begin{figure*}[t]
  \centering
  \includegraphics[width=\textwidth, trim=0 230 0 0, clip]{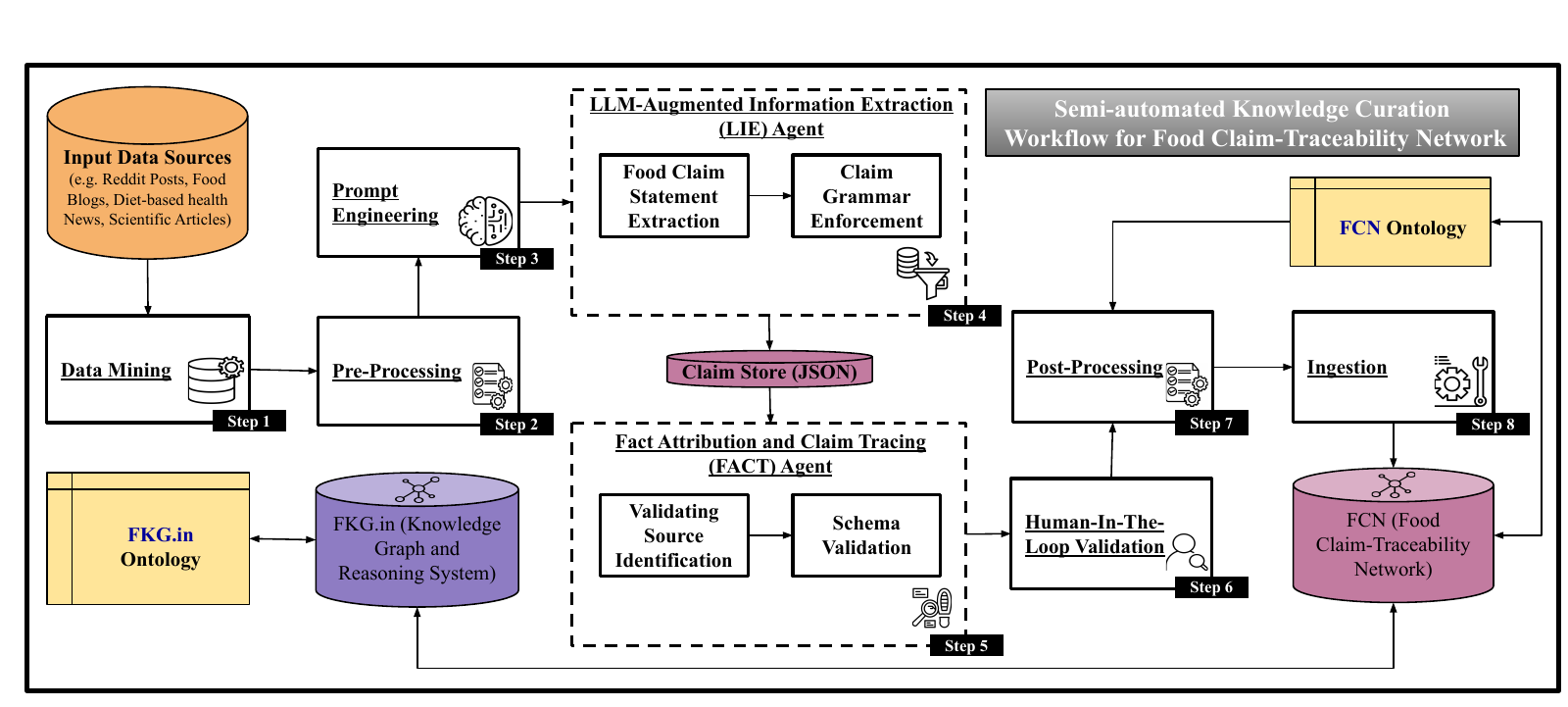}
  \caption{Semi-Automated Knowledge Curation Workflow for FKG.in-FCN} 
  \label{fig:knowledge_curation_workflow}
  \Description{Overview of the semi-automated knowledge curation workflow for FKG.in. The pipeline ingests food-related texts from diverse sources, extracts candidate claims using LLM-assisted parsing, and links them to structured entities in the Food Claim Ontology. Validation sources are retrieved or annotated, and claims are enriched with metadata such as stance, context, and modality. The workflow balances automation and expert oversight to support scalable, provenance-rich modeling of food claims across domains.}
\end{figure*}

\par The Food Claim Network (FCN) extends the FKG.in ontology to represent individual food claims as structured, traceable entities. The workflow semi-automates extraction, normalization, validation, and knowledge graph ingestion of claims from heterogeneous sources. It integrates LLM-based information extraction, schema enforcement, and human-in-the-loop validation to build a scalable, high-quality claim-level knowledge infrastructure.

\par We begin by curating a diverse set of food-related sources, from informal user-generated content (e.g., Reddit) and food blogs to semi-formal diet-focused news sites and formal scientific literature (e.g., PubMed). This heterogeneity reflects the range of claim origins in the food discourse space. Sources are categorized by type, credibility, and linguistic format to ensure breadth and contextual depth. This foundational step is critical for representing how food claims propagate across scientific, cultural, and popular knowledge systems. The workflow components in figure \ref{fig:knowledge_curation_workflow} are described below:

\begin{itemize}
    \item \textbf{Step 1: Data Mining} - Raw text is extracted via web scraping, API retrieval, or open data dumps (e.g., Academic Torrents). Domain-specific filters identify content likely to contain food-health claims using keyword heuristics (e.g., "boosts immunity," "good for digestion"), URL patterns (for blogs or scientific journals), and metadata tags. Mined text is stored in a standardized format for uniform downstream processing.
    \item \textbf{Step 2: Pre-Processing} - Text is cleaned and standardized, including stripping HTML, removing social media artifacts, collapsing whitespace, and segmenting passages into sentences or fragments. Candidate food mentions are detected using lightweight NER models or lexicons. This prepares discrete, analyzable units and improves recall of food-related statements for the extraction phase.
    \item \textbf{Step 3: Prompt Engineering} - To extract claims reliably with LLMs, we design few-shot prompts that guide the model to generate structured claim statements along with their source, context, and validation counterparts from the preprocessed text, focusing on atomicity, relevance, and fidelity to the original source and language. Outputs are schema-conformant, distinguishing the food item, grammatical segments of \textbf{Food Claim} and \textbf{Validating Source}, and contextual qualifiers. Prompt design is iterative and informed by ontology constraints and empirical testing. OpenAI's GPT-3.5 Turbo \footnote{https://platform.openai.com/docs/models} is currently used in this pipeline.
    \item \textbf{Step 4: LLM-Augmented Information Extraction (LIE) Agent} - The LIE Agent is the workflow's core and has two submodules. \textbf{Food Claim Statement Extraction} uses engineered LLM prompts to extract claims strictly grounded in the input text, decomposing compound assertions into atomic claims centered on a single food item. \textbf{Claim Grammar Enforcement} restructures each claim into a consistent schema - typically subject (food) + property + effect - with optional fields like mechanism, condition, etc., as described in Subsection \ref{subsec:ontology_design}. This two-stage process ensures both fidelity and structural integrity of claims. We deliberately (and ironically) call it the "LIE" Agent because, at this stage, the extracted information reflects the provisional nature of claims, which are unverified at this stage. Claims are stored in structured JSON along with metadata like source URL, retrieval date, and raw text snippet, in addition to the parsed claim fields. This modular representation supports integration into a knowledge graph, version tracking, analytics, and semantic search by food entity, claim/effect type, or provenance.
    \item \textbf{Step 5: Fact Attribution and Claim Tracing (FACT) Agent} - The FACT Agent has two main functions. \textbf{Validating Source Identification} uses rule-based extraction or LLM prompts to locate and parse supporting or refuting sources from the original text or related discussions (e.g., Reddit counterclaims, inline citations, hyperlinks), tagging each with stance, confidence, source type, and organization. \textbf{Schema Validation} ensures these validations conform to the defined schema, linking correctly to claims, filling missing fields using default values, flagging malformed entries, and enforcing consistency (e.g., an effect must have a subject). We call it the "FACT" Agent, not to imply ful verification, but to denote the stage where attribution, counter-evidence, and structural validation start, moving the claim closer to factual grounding without guaranteeing its truth.
    \item \textbf{Step 6: Human-in-the-Loop Validation} - Despite automation's scalability, human oversight remains essential. Domain experts or trained annotators review a subset of extracted claims to assess quality, verify ambiguous mappings, and suggest schema improvements. This feedback calibrates prompts, reduces hallucinations, and increases trust. Future iterations aim for active learning setups prioritizing uncertain outputs for human review and an interactive annotation platform for direct claim inspection and correction.
    \item \textbf{Step 7: Post-Processing} - Extracted and validated claims are prepared for final ingestion into Food Claim-traceability Network (FCN) graph. This step includes deduplication, entity normalization (e.g., "skincare" vs "skin" as a claim effect type), enforcing consistent identifiers, and semantic enrichment (e.g., linking "digestion" to controlled vocabularies). Optional auxiliary metadata such as region, cultural context, or language can also be integrated for richer analysis. 
    \item \textbf{Step 8: Ingestion in the FCN Knowledge Graph} - Processed claims are converted into RDF triples and ingested into FCN. As an extension of FKG.in, FCN is designed specifically to capture claim structures, stance indicators, and validation metadata. Each graph entity (claim, source, food item, effect, validating source, etc.) gets a persistent identifier for long-term traceability and interlinking. To enhance interoperability, we link FCN entities to corresponding FKG.in entities. E.g., a "turmeric improving digestion" claim links the turmeric node in FCN to its canonical FKG.in representation, enabling semantic inheritance, integrated search, and analysis alongside compositional, cultural, and sensory data.
\end{itemize}

\subsection{Proof of Concept}
\label{subsec:proof-of-concept}

\par To demonstrate the feasibility of our end-to-end pipeline - from data collection and claim detection to entity linking and validation - we conducted a preliminary run on a controlled subset of user-generated content. Reddit was chosen for its informal, community-driven food discussions. We applied our system to curated long-form posts and confirmed that the pipeline could parse free-form text, identify food claims, extract entities, and associate them with validation contexts. This small-scale run validated the modular design and functional coherence of our workflow and demonstrated the potential to structure unregulated food discourse into machine-readable knowledge. Results are discussed in the next section.

\section{Results and Discussion}
\label{sec:results}

\par
\begin{figure*}[t]
  \centering
  \includegraphics[width=\textwidth]{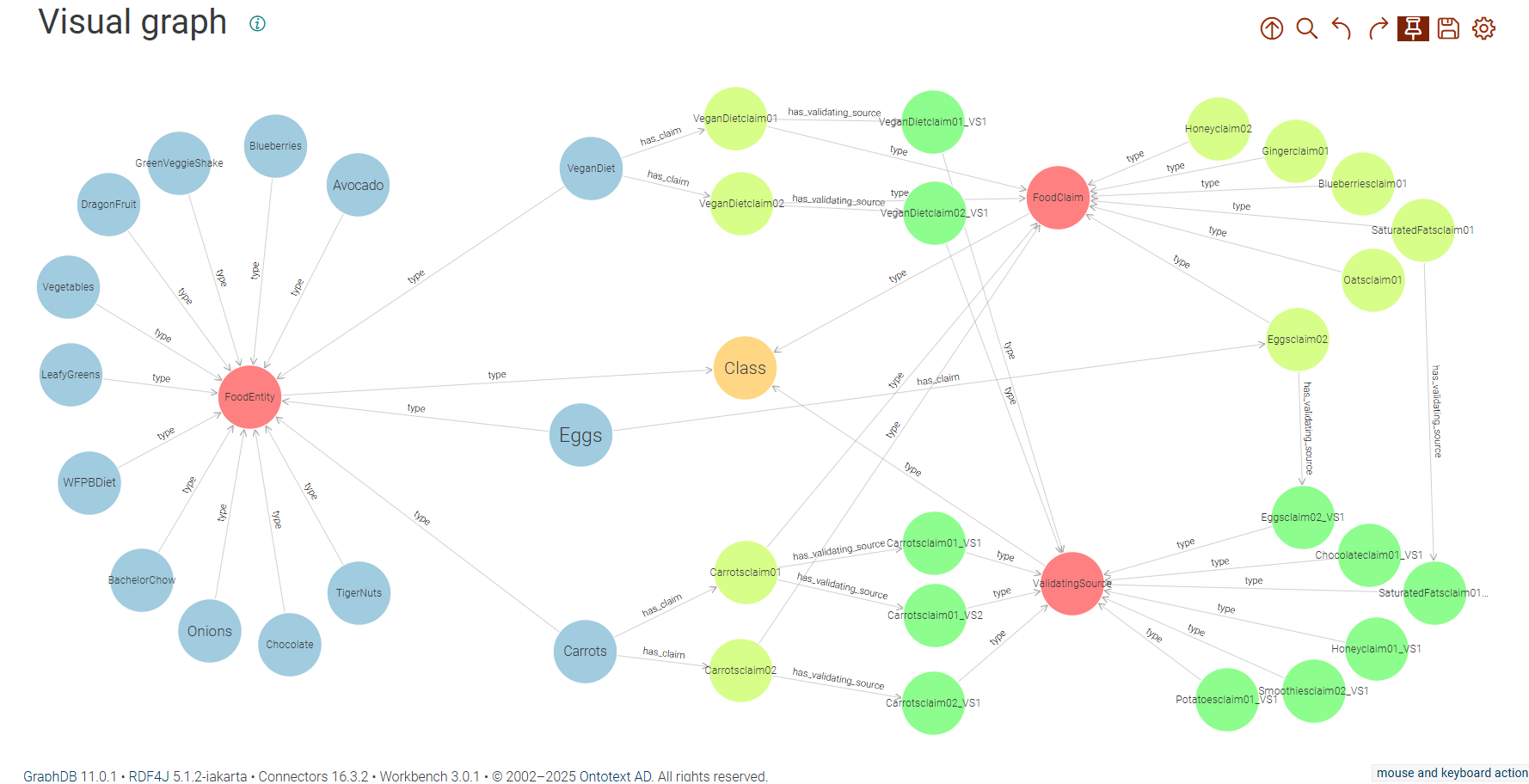}
  \caption{Sample FCN Knowledge Graph (GraphDB) with Food Entity, Food Claim, and Validating Source categories} 
  \label{fig:knowledge_graph_sample}
  \Description{This composite figure illustrates the structure and composition of the FCN Knowledge Graph, hosted on GraphDB, with a focus on three primary classes: FoodEntity, FoodClaim, and ValidatingSource.}
\end{figure*}

 A multi-stage AI-augmented pipeline analyzed a curated corpus of 515 long-form Reddit posts (each >1,500 characters) on health and food. Only 89 posts (17.3\%) met our criteria for containing food-related claims. From these, 187 distinct entities (food\_entities) were extracted, mapping to 265 unique subjects (claim\_subject) because a single post may contain claims about multiple subjects despite the post being about one food entity. Totally, 459 claims (claim\_text) were identified, alongside 578 associated texts providing supporting or debunking evidence (validity\_text). The resulting dataset was visualized as a semantic graph in GraphDB (Figure \ref{fig:knowledge_graph_sample}), to illustrate entity-claim relationships and validation links. The full graph has 6,592 nodes and \textasciitilde 3,000 \textbf{Food Claim}, \textasciitilde 2,000 \textbf{Validating Source}, and \textasciitilde 1,000 \textbf{Food Entity} links, demonstrating a dense, interlinked knowledge structure suitable for reasoning and analysis. Manual validation compared the system's JSON output against the original posts. The review confirmed high efficacy: 28 unique food/dietary entities were identified, and 46 of 58 potential health claims were extracted (79.3\% recall). Sample performance on complex tasks was notable with a 100\% success rate linking all 12 cited URLs. Weaknesses included "claim splitting," where the model captures a paragraph's primary claim but misses secondary claims (accounting for all 12 missed claims), and entity resolution errors, where distinct subjects were grouped under a generic entity. These findings highlight further need for prompt refinement.

\par Sample data for all the \textbf{Food Entity}, \textbf{Food Claim}, and \textbf{Validating Source}, as produced on GraphDB, along with exploratory data analysis (EDA) and word cloud analyses performed on FKG.in-FCN are available in the project repository\footref{fn:repo} and archived on Zenodo\cite{fkg-india_foodcomputing_resources}. The EDA reveals key patterns in food-health discussions on Reddit. The distribution of effect types for the top 10 claim subjects shows that foods like Aronia berries and cardamom span diverse health effects, reflecting versatility. Additionally, most claims center on cardiovascular health, while diets like WFPB (Whole-Food, Plant-Based) and Keto are linked to weight, sleep, and mental health, reflecting lifestyle concerns. Top individual effect types reveal that "health" (217 mentions), "metabolism" (47), and "digestion" (29) dominate, suggesting foods' general association with systemic benefits. Grouped into broader effect categories, "General Health \& Longevity" is most cited (212 claims), followed by "Metabolism \& Energy" and "Diet \& Weight Management," highlighting focus on holistic and lifestyle outcomes. Analysis of validating sources shows "scientific studies" and "expert quotes" to be most cited, though "anecdotes" and "testimonies" also appear frequently, reflecting a mix of formal and informal reasoning of food-health claims. This demonstrates the influence of scientific authority alongside the persuasive power of personal experience in online dietary discourse.

\section{Limitations and Future Directions}
\label{sec:limitations}

\par While our semi-automated pipeline demonstrates the feasibility of food claim extraction and traceability, several limitations remain. Prompt engineering for LLMs is brittle, as small phrasing changes can drastically affect output, particularly with informal, ambiguous, or culturally situated text. Few-shot prompts require iterative tuning, limiting scalability. Claims spanning multiple foods, implicit cultural context, or unclear syntax are difficult to decompose into atomic, schema-conformant assertions. Source attribution is often incomplete due to missing metadata or subtle wording shifts. Human-in-the-loop review helps, but community participation, multilingual parsing, and modality expansion remain open challenges. Also, scaling human verification across languages and cultural contexts requires substantial resources, time, and expertise. Finally, as with all knowledge infrastructures, bias, authority, and epistemic pluralism must be addressed to avoid flattening cultural nuance or reifying dominant narratives.

\par Future work will expand both the depth and breadth of the FCN. Depth-wise, we aim to add multilingual capabilities to capture claims expressed in regional Indian languages and enable culturally-aware semantic normalization. Breadth-wise, we plan to include non-textual modalities such as packaging imagery, voice-based folk knowledge, and marketing videos - using OCR, speech-to-text, and multimodal embeddings. We also envision user-facing interfaces that allow consumers, researchers, and policymakers to search, flag, and contextualize claims within a live, growing knowledge graph. To strengthen credibility and cross-verification, we will incorporate Retrieval-Augmented Generation (RAG) agents drawing on trusted corpora such as scientific literature and regulatory archives. Automated validation metrics will be developed to track extraction fidelity, claim completeness, and entity resolution accuracy. We also plan to link FCN with nutrition calculation pipelines, medical literature indices, regulatory notices, and food safety data, some of which are being developed actively. Ultimately, we aim to evolve FCN into a collaborative platform where community contributions, expert review, and AI systems co-evolve into a dynamic, traceable, and trustworthy food information ecosystem.

\section{Conclusion}
\label{sec:conclusion}

\par The Food Claim-Traceability Network marks a critical step toward making food discourse transparent, accountable, and computable. By modeling food claims as structured entities - anchored in provenance, modality, and context - we move beyond static fact-checking to support a dynamic understanding of how food knowledge is produced, circulated, and contested. Our proof-of-concept with Reddit data leverages LLM-augmented, semi-automated, and ontology-driven workflow to demonstrate the feasibility of tracing food claims at scale and lays the groundwork for a food knowledge infrastructure supporting misinformation detection, claim validation, and public health interventions. In doing so, we offer not just a technical framework, but a conceptual shift: from decontextualized statements to semantically rich ecosystems of meaning. 

\par What emerges is a framework for epistemic pluralism, accommodating diverse forms of knowledge (scientific, traditional, commercial), participatory validation, and adaptation to evolving food discourse. While focused on India, the framework is globally extensible, with potential applications in food regulation, digital health, consumer education, and AI safety. In an era of nutritional confusion, polarized media, and AI-generated content, FCN lays the foundation for an auditable, interpretable, and equitable approach to understanding food claims. By tracing their lifecycle across cultures, time and media, we aim to debunk myths, preserve context, amplify credibility, and democratize access to food knowledge, thus empowering communities to understand what is claimed about food, where it comes from, and why it matters.

\begin{acks}
\par This research was supported by the Mphasis AI \& Applied Tech Lab at Ashoka - a collaboration between Ashoka University and Mphasis Limited.
\end{acks}

\bibliographystyle{ACM-Reference-Format}
\bibliography{main}


\end{document}